%% file: main.tex
\documentclass[runningheads]{llncs}
\usepackage{graphicx}
\usepackage{fullpage}
\usepackage{textcomp}
\usepackage{amssymb}
\usepackage{booktabs}
\usepackage{float}
\usepackage{times}
\usepackage{latexsym}
\usepackage{graphicx}
\usepackage{tabularx}
\usepackage{url}
\usepackage[inline]{enumitem}
\usepackage{array}
\usepackage{setspace}
\usepackage{multirow}
\usepackage{multicol}
\usepackage{hyperref}
\usepackage{framed}
\usepackage{color}
\usepackage{xcolor}
\usepackage{xspace}
\usepackage{amsmath}
\usepackage{amssymb}
\usepackage{arydshln}
\usepackage{makecell}
\usepackage{amsmath}
\usepackage{xcolor}
\usepackage[inline]{enumitem}
\usepackage[caption=false,font=footnotesize]{subfig}
\usepackage{stfloats}

\newtheorem{assumption}{Assumption}

\newcommand{\ubuntudataset}{UbuntuDataset}
\newcommand{\malwaredataset}{MalwareDataset}

\newcommand{\transformer}{Transformer}
\newcommand{\seqseq}{Seq2Seq}
\newcommand{\debin}{DEBIN}
\newcommand{\transformerPT}{TransformerPT}
\newcommand{\scratch}{TFScratch}

\begin{document}
\title{In Nomine Function:  Naming Functions in Stripped Binaries with Neural Networks}

\titlerunning{In Nomine Function}
%
\author{Fiorella Artuso$^1$ \and Giuseppe Di Luna$^2$ \and Luca Massarelli$^2$ \and Leonardo Querzoni$^2$}
\authorrunning{Artuso et al.}
%
\institute{CINI, Italy. 
\email{fiorella.artuso23@gmail.com}\\
\and
Sapienza, University of Rome, Italy. \\
\email{\{diluna,massarelli,querzoni\}@diag.uniroma1.it}}
\maketitle              
\begin{abstract}
In this paper we investigate the problem of automatically naming pieces of assembly code. Where by naming we mean assigning to an assembly function a string of words that would likely be assigned by a human reverse engineer. 
We formally and precisely define the framework in which our investigation takes place. That is we define the problem, we provide reasonable justifications for the choices that we made for the design of training and the tests. We performed an
analysis on a large real-world corpora constituted by nearly 9 millions of functions taken from more than 22k softwares. In such framework we test baselines
coming from the field of Natural Language Processing (e.g., Seq2Seq networks and Transformer). 
Interestingly, our evaluation shows promising results beating the state-of-the-art and reaching good performance. We investigate the applicability of tine-tuning (i.e., taking a model already  trained on a large generic corpora and retraining it for a specific task). Such technique is popular and well-known in the NLP field. 
Our results confirm that fine-tuning is effective even when neural networks are applied to binaries. We show that a model, pre-trained on the aforementioned corpora, when fine-tuned has higher performances on specific domains (such as predicting names in system utilites, malware, etc).

\end{abstract}

\input{sec_1_introduction.tex}
\input{sec_2_related_work.tex}
\input{sec_3_problem_definition.tex}
\input{sec_4_dataset.tex}

\input{sec_5_solution_overview.tex}

\input{sec_6_exp_result.tex}

\input{sec_7_human.tex}

\input{sec_8_conclusion.tex}

\bibliographystyle{plain}

\end{document}

%% file: sec_1_introduction.tex
\section{Introduction}

Last few years have witnessed the growth of a trend consisting in the application of machine learning (ML) and natural language processing (NLP) techniques to the code analysis field, as illustrated in \cite{sourcecode}. In fact, the vast and increasing number of high quality software available through open source repositories such as GitHub, has given the chance to leverage large amount of source code as a ground truth for building statistical models of code. The design choice of using NLP to build such models is motivated by the naturalness hypothesis which underlines the similarities between programming languages and human languages. According to this hypothesis, software is a form of human communication with similar statistical properties to natural language and these properties can be exploited to build better software engineering tools \cite{sourcecode}. The practice of applying ML and NLP techniques to code turned out to be very helpful and effective in many tasks such as predicting program bugs \cite{defect_prediction}, predicting identifier names \cite{context2name}, translating code between programming languages \cite{chakraborty2018tree2tree}, etc. Thus, the success given by the application of NLP techniques to source code has led to investigate the possible use of such techniques also in the context of binary code analysis; learning function signatures \cite{signature}, identifying similar functions \cite{ASM2VEC,SAFE,GEMINI}, recovering the compiler that generated a given binary \cite{BAR19}, just to cite a few.

Following this research line, in this paper we investigate the feasibility of using similar techniques to predict the name of functions in stripped binary programs. The latter are binary executable files that only contain low-level information such as instructions, registers and memory addresses but no debug symbols since they are not directly necessary for program execution. Debug symbols are generated by compiler programs on the basis of the source code and typically include information about functions and variables, such as name, location, type and size which are helpful for debugging and security analysis of a binary. Being non essential for the software execution, symbols are often removed from a program after compilation, increasing the complexity of reverse engineer the software.

Reconstructing symbols in a binary program can be a very useful feature for all those field where reverse engineering code plays a crucial role, e.g. malware analysis. Usually, after having disassembled a malware the reverse engineer starts analyzing the set of assembly instructions of the program looking for  specific functions (e.g. encryption or network) that might reveal the malicious nature of the software sample under investigation. This task could be daunting, especially when the binary code is stripped and original function names are not present. In this case, it could be very helpful having explanatory names for such functions, as they would save a lot of effort to the reverser whose work could be guided and supported by clear hints about the content of each function.

Recently, a few works proposed solutions to this problem. For example, \cite{DEBIN} and \cite{NERO} have shown promising results in predicting functions names. However, the problem at hand is far from being solved as existing solutions work only under strong assumptions, like the presence of symbolic calls to dynamically linked libraries in each functions \cite{NERO}, or a closed set of possible assignable names to predict from \cite{DEBIN}. Furthermore, existing solutions have only be evaluated on small datasets that contain binary programs with little variance (less than 1k softwares), and this datasets have not been made publicly available, hindering the possibility to compare existing solutions.

Starting from these issues in this paper we propose the following contributions:

\begin{itemize}
	\item {\bf Dataset.} We created a new dataset, namely \ubuntudataset, composed by a large number of real world binaries, spannig several heterogeneous application fields: networking, databases, video-games, system utilities, c libraries, etc. Our dataset contains $8.8$ millions functions from 22k distinct softwares. The dataset is made available online to the community for testing and evaluation.
	
	\item {\bf Competitive DNN solutions.} Considering the function naming problem, we train and test two Deep Neural Network architectures \seqseq{} and \transformer{}.  We achieve an f1-score of $0.122$ for \seqseq{} and of $0.230$ for \transformer{}. We compare our architectures with previous solutions on a subset of our test set, showing an improvement with respect to the state-of-the-art. 
	
	\item {\bf Fine-tuning.}  We show that a model trained on our \ubuntudataset{} has a drastic performance improvement when fine-tuned for a specific domain.  Fine-tuning \cite{BERT} is a popular technique in NLP, in which models trained on huge general purpose datasets are then re-trained on a smaller dataset specialised for a certain sub-task. In this paper we show that fine-tuning works also when applied on models for binary code: we found a marked improvement of our model when fine-tuned to name applications of a specific domain (such as system utilities). In particular, we show that a model fine-tuned on \emph{busybox} has a relative improvement of 22\% when tested on \emph{coreutils} with respect to a pre-trained model with no fine-tuning. We also show that fine-tuning helps the network in having better performances when naming functions across different optimisations. That is, we fine-tune our network on certain set of packages compiled with optimisation flag O$0$ and then we show that this leads to a two-fold improvement when naming the same packages compiled with O$1, $O$2$, and O$3$. We also compare all our fine-tuned models  to models that are trained from scratch only on the specific fine-tuning dataset. As expected, these latter models have lower performances. This shows that fine-tuning is able to exploit and refine the knowledge learnt on large general datasets of binaries, but the specific training sets used for this purpose do not contain enough information to learn from scratch the correct relationships name-assembly code. 

	\item {\bf Test on Malware.} We test our solution on real world linux malware in a quantitative and qualitative way. Also in this case we test performances with and without fine-tuning, showing a 5-fold performance improvement in the former case.
\end{itemize}



After this introduction, Section \ref{sec:related_work} discusses the state of the art, Section \ref{sec:problem} introduces the function naming problem, Section \ref{sec:dataset} describes the \ubuntudataset{} used for the evaluation, Section \ref{sec:solution} details the proposed solutions.  Section \ref{sec:experiments} reports the experimental results on goodware, and Section \ref{sec:malware} reports our tests on malware. Finally, Section \ref{sec:conclusion} concludes the paper.

%% file: sec_2_related_work.tex
\section{Related Work}
\label{sec:related_work}

There are several works on predicting names for variables, functions and objects using statistical learning. 

\subsection{Prediction of Variables, and Function Names in High Level Languages}
A body of works have explored the possibility of predicting  variable, function and object names within code expressed in high-level programming languages (such as Java, Javascript, C and similars), the majority of which leverage deep learning techniques.
In \cite{classname} a word2vec-like approach is used to learn a probability distribution on class and method names. The model can then be used to suggest probable names for classes, methods and variables in Java source code.
In \cite{AllamanisPS16} convolutional neural networks are used to create extremely summarised descriptions of Java functions that resemble function names.
Code2seq \cite{code2seq} proposes an encoder-decoder strategy on the Abstract Syntax Tree (AST) of functions to obtain explanative sentences, so called ``code captions'', of C\# functions and Java methods. 
Finally, \cite{j2019} uses conditional random fields to predict names of variables and methods in large Javascript programs. 

This line of works leverages the huge amount of information that is available with high level code (e.g. variable types, other names, etc.), information that are not present in compiled binary code. Therefore, directly applying these methodologies to our work is unfeasible. 

\subsection{Prediction of Variables, and Function Names in Binary Code}
Predicting debugging symbols, including function names, through ML is a rather new field of research with few contributions. The most notable work in this area is DEBIN, proposed by He et al.~\cite{DEBIN}. It uses conditional random fields to predict debug symbols. similarly to \cite{j2019} it also predicts function names, variable names and types. Differently from our work, DEBIN is only able to assign to functions names from a predetermined closed set, i.e. it cannot generalise to new names. The dataset used in ~\cite{DEBIN} to evaluate DEBIN is composed by 9k executables from 830 linux packages. 

DEBIN's limitation is surpassed by a recent pre-print, NERO~\cite{NERO}, which models the problem of predicting function names as a Neural Machine Translation (NMT) task where each function is represented by ``call sites sequences'': each call site is an encoding of a function call containing the name of the called function and information on the parameters. Unfortunately, it is not clear how NERO will cope with functions that do not use calls, since it seems that no feature is extracted in this case. Additionally, by taking into consideration only calls to external library, their method is likely to miss the difference functions that rely on the same set of api calls but are constituted by markedly patterns of instructions (e.g., encryption functions and sorting functions implemented from scratch). 

Finally, DIRE~\cite{lacomis2019dire} proposes a probabilistic technique for variables name recovery that uses both lexical and structural information
recovered by using a decompiler. DIRE uses the decompiler’s internal AST representation, which encodes additional structural information, to train a neural network based on an encoder-decoder structure, where the
encoder is composed by a bidirectional-LSTM and a graph encoder and the decoder is a standard LSTM. Their entire dataset contains $1259$k functions, with a test set of $120$k functions. We remark that recovering variables name is a different task than ours, and that function names are used in DIRE to help in predicting the name of variables.

%% file: sec_3_problem_definition.tex
\section{Framework and Problem definition}

\label{sec:problem}
\subsection{Problem Definition}
Given a fragment of binary code $b$ representing a functional unit of code (for simplicity, and without loss of generality, we assume it corresponds to a function) in a compiled software, we want to output a string $s$. Such string $s$ has to represent a ``{\em meaningful}" name for function $b$, that is a name that captures its semantic and its role inside the software. 

The above problem is extremely challenging, and, due to its nature, cannot be defined more precisely without incurring in complex reasoning about what is the ``semantic" of code. Fortunately, statistical learning methods are especially suitable for problems where the definition itself is fuzzy.  As customary in statistical approaches we will try to learn a probability distribution that assigns to each function $b$ the most probable output string $s$ using a large dataset of assembly functions with semantically expressive names. 

Our investigation of the function naming problem is based on a set of simplifying assumptions that made the training and testing phase tractable. In the following we precisely state all our assumptions, providing the reasoning behind our design, and we describe the main challenges of the problem. 

\subsection{Training} 
Statistical methods, especially the ones based on neural networks, are effective when trained on a suitable dataset of relatively large size (millions of functions). In a dataset that counts millions of functions it would be unreasonable to manually annotate each one with a representative name. Therefore, in this paper we assume the following:

\begin{assumption}{(Sensible programmer)} 
A programmer, when writing code in an high-level language, assigns names to functions that represent their semantics and their role inside the software.
\end{assumption}

The above assumption does not always hold true. However, we find reasonable to assume that it holds most of the time, especially, in large projects developed by professional or skilled programmers and where common naming conventions are often used and enforced. This assumption allows us to create a dataset by disassembling the binary code from open source projects, without the need to perform manual annotation as function names are available in the source code. 

\subsection{Testing}\label{challange:testing}
There is an unavoidable ambiguity in the output of any method that names something: in general several different meaningful descriptive names can be associated to a given piece of code. As an example, a function implementing quick sort on an array could be named ``Quick-sort", ``quick-array-sort", ``sort", etc. This creates a problem in the way our solution is tested: in order to evaluate its real accuracy we should consider all possible meaningful names, but this is, again, unfeasible.

Therefore, as common in the NLP literature \cite{bahdanau2014neural,NovikovaDCR17}, we will say that a prediction is correct if it is the same name present in the dataset. In our case we will measure the performance using the classical metrics of precision, recall and F1. The drawback of this evaluation methodology is that other predictions will be deemed as wrong even if they are meaningful.  

\subsection{Vocabulary and Restricted Names}
Unfortunately, we found that functions names in our dataset are noisy, as example many functions of the OpenGL library contain the bigram ``gl''. Such pattern is recurrent in many libraries and softwares; this is due to the fact that developers use words and acronyms that have the meaning of identifying the software itself, but that do not add much to the semantic of the function. In order to clean our dataset we designed a filtering process (described in a detailed way in Section \ref{sec:dataset}). This filtering process has the purpose of associating each original name to a reduced name over a restricted vocabulary of words (whose size was around 1k words in our experiments). We found that such restriction preserves the semantic of the majority of names in our dataset and it solves the problem described above. 
%

%% file: sec_4_dataset.tex
\section{The \ubuntudataset}
\label{sec:dataset}
In this section we describe our dataset, and the steps used for its construction. The dataset can be found at the following URL  \url{https://github.com/gadiluna/in_nomine_function}.

\subsection{Dataset Building Process}

\ubuntudataset{} was built by downloading all available amd64 packages from the Ubuntu 19.04 apt repositories \footnote{Namely, \emph{main}, \emph{restricted}, \emph{universe} and \emph{multiverse}.}. We collected a total of 22040 packages; for each package, when available, we downloaded the corresponding debug symbols and extracted all the executables files.
At the end of this process we got 87853 distinct ELF files that we disassembled with IDA Pro. We then filtered out duplicated functions\footnote{Two functions are duplicate if they contain the same list of instructions after the substitution of constants and memory addresses with a dummy value.} and we discarded functions for which we failed to get a symbolic name in the debug symbols repository. At the end of this process we got 10005866 distinct named functions. These functions have been additionally filtered by a process described below, and at the end we obtained  nearly 9 millions usable functions that constitute our Training, Validation and Test dataset. Table \ref{tab:datasetcomp} reports a comparison among datasets used by papers that address (directly or indirectly) the function naming problem on binaries.

\begin{table}[]
\caption{Comparison of datasets used by function naming papers. {\bf (**)} The paper does not report the total number of functions, but it reports an average of 79 functions per binary file and a total number of 9000 binaries. {\bf (*)}: Nero pre-print \cite{NERO} mentions a public dataset but no link to it is provided.}
\center
\begin{tabular}{|l|l|l|l|l|}
\hline
\textbf{Paper} & \textbf{\#softwares} & \textbf{\#functions} & \textbf{Operating system}         & \textbf{Released} \\ \hline \hline
\ubuntudataset{} (Our)           & 22040                & $8861407$                  & Ubuntu 19.04 Repositories  & yes               \\ \hline
Debin \cite{DEBIN}   & 830                  &    711k  Ext. {\bf (**)}                    & Ubuntu                  & No              \\ \hline
Nero \cite{NERO}    &             Unknown         & 15k                       & Linux  Generic                       & No {\bf (*)}          \\ \hline
\end{tabular}
\label{tab:datasetcomp}

\end{table}

\subsection{Function Representation}

In \ubuntudataset{} each function is represented by its linear list of instructions. The relative order in the list is given by the address of each instruction inside the program. The average number of instructions per function is $159.12$; we truncated all functions with more than $500$ instructions and we removed all functions with less than $5$ instructions.  

\subsection{From Function Names to Tokens}\label{sec:normname}

We apply a normalization process to function names whose final goal is to represent each of them as a list of tokens in a reference vocabulary. The process of normalisation is based on six steps:
\begin{enumerate}
	\item {\em Demangling}
	\item {\em Splitting}
	\item {\em Stemming}
	\item {\em Vocabulary construction}
	\item {\em Final conversion}
\end{enumerate}

\paragraph{Demangling.} The names contained in the debug symbol table have been mangled by the compiler for various reasons (e.g, implementing methods overriding). The aim of this first step is to recover the original function names. Since compilers perform mangling in a standard way, it is possible to perform name demangling using standard libraries \footnote{In particular, we used cxxfilt:{\url{https://github.com/afq984/python-cxxfilt }}}. During this step we filtered all the binary functions deriving from source code written in languages different from C/C++ (e.g. GO, Haskell, etc.). 

\paragraph{Splitting.} This step consists in splitting function names into tokens. This is achieved by using the natural partition provided by \emph{camelcase} and \emph{snake} notations, which are generally adopted for function names.

\paragraph{Stemming.} This is a technique used in information retrieval to reduce inflected (or sometimes derived) words to their base form. Nevertheless, such technique turns out to be very useful in this context, since it has the effect of reducing the vocabulary  by mapping different forms of the same token into a unique one (for example the tokens ``shared'' and ``sharing'' are both mapped to their base form ``share''). Stemming is important since it is not necessary for the network to learn the syntactically correct token to associate to each function, but rather the semantically correct one.

\paragraph{Vocabulary construction.} This step consists in creating a meaningful vocabulary of tokens. This operation has the goal of removing useless and meaningless tokens from function names (e.g., many functions in the gnu scientific library starts with the token ``gsl"). The token selection process consists first in assigning a score to each token and then retaining only tokens whose score is above a certain threshold $\tau$.
The score for each token $t$ is the project frequency, that is the number of different packages in which a token appears. This choice permit us to exclude tokens that appear only in few packages independently from the frequency of the token inside the package. In this way we avoid to assign a large score to tokens that are not semantically relevant (see the aforementioned example of the ``gsl" token).
We set $\tau$ to 500 obtaining 1080 tokens. Finally, we exclude from the vocabulary all token that have no meaning ending up with a vocabulary of 1064 tokens. 

\paragraph{Final conversion.} This final step consists in removing from function names all those tokens not contained in the vocabulary built in the previous step. Moreover, this step is also designed to split function names that are not using the camelcase or snake notation. The idea is to split tokens in case they match a word in our vocabulary, as example the token ``numpy" matches the vocabulary word ``num", therefore we transform it in the token ``num"; the token ``numvertex" is transformed in ``num" and ``vertex". There are few cases in which function names only consist of out-of-vocabulary tokens resulting in an empty name; in such cases functions are removed from the dataset.  At the end of the entire normalisation process we  filtered out 11\% of the initial functions and we obtained nearly 9 millions of functions. 

\paragraph{Statistical analysis of function names.}
\begin{figure*}[t]

\subfloat[Frequency of the 25 most frequent tokens.  \label{fig:top15names}]{\includegraphics[width=.5\textwidth]{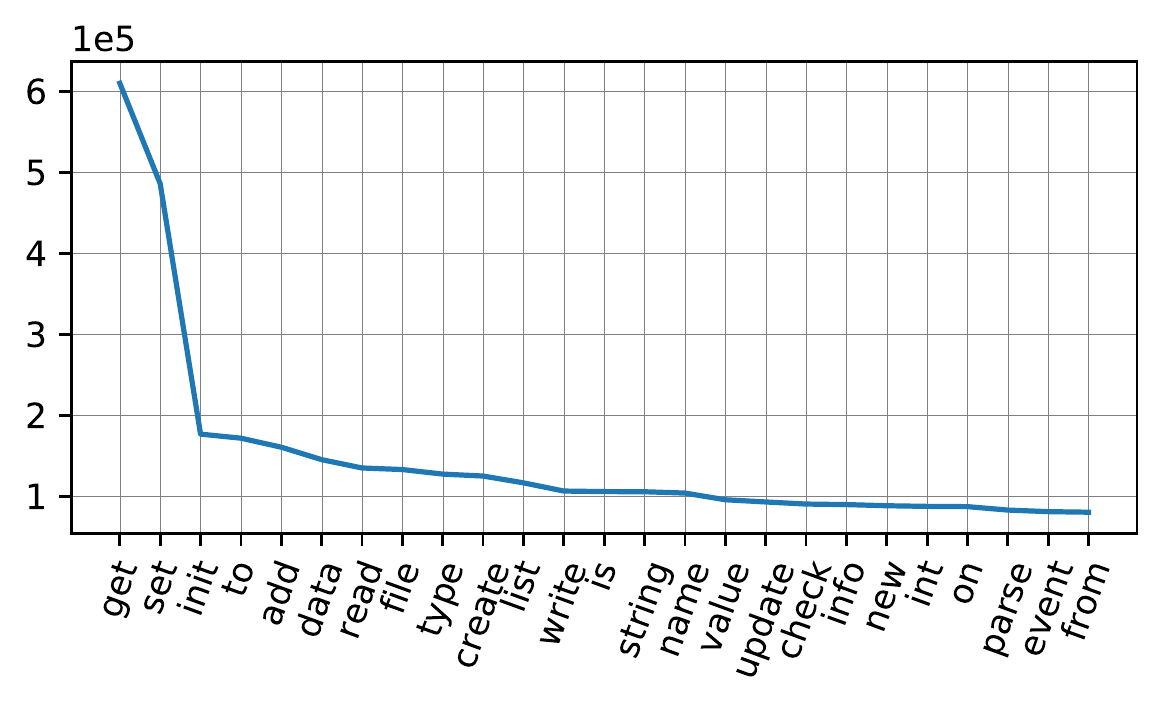}}
\subfloat[Log-log scale plot of the frequency of all tokens in our vocabulary.  \label{fig:loglog} ]{\includegraphics[width=.5\textwidth]{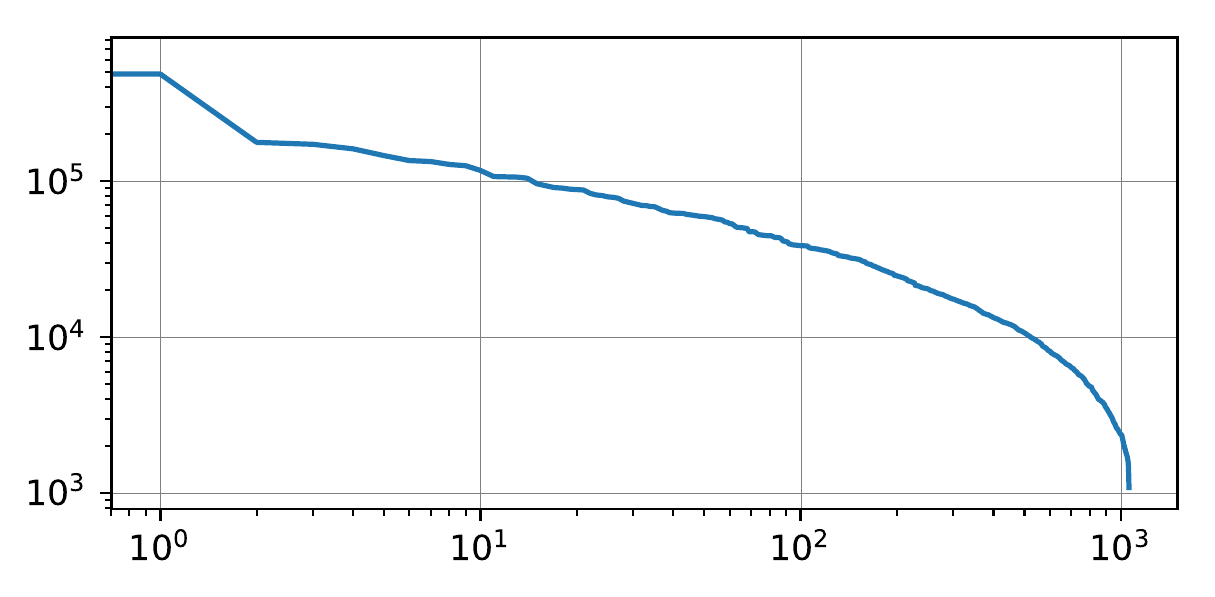}}
\caption{Statistical analysis of the vocabulary.}\label{fig:analysisfreq}
\end{figure*}
We performed some basic analysis on the names contained in \ubuntudataset{}. 
The average numbers of tokens in a name is $2.7$. 
The $98\%$ of all names is composed by  $8$ tokens or less. The frequency distribution of the 25 most frequent token is in Figure \ref{fig:top15names}, while in Figure \ref{fig:loglog} we report the log-log plot of the frequency of all tokens in our vocabulary. 
It is not surprising that get and set are predominant tokens, this is due to their frequent use in objected oriented programming paradigm. Furthermore, most of the words are likely distributed according to a power law distribution, this is consistent with the distribution of words in human languages (as example the english words seems to follow a Zipf-law \cite{plos17}). However, such fit is not perfect: in the log-log plot the tails of the distribution  (the highest and lowest frequency words) do not fit a linear interpolation. 
\vspace{-0.3cm}
\subsection{Train, Validation, and Test Splits}

In order to train and evaluate the models we split the dataset in the canonical Train, Validation and Test sets. We put $80\%$ of functions in the Training set, $10\%$ in the Validation set and $10\%$ in the Test set (ie. Train set size: $7064820$,Validation set size: $944263$, Test set size: $852324$). We avoid information leakage by splitting functions by packages: all functions from a given package belong to only one of the three sets.

%% file: sec_5_solution_overview.tex
\section{Solution Overview}
\label{sec:solution}
In this section we describe the solutions we tested. We considered two different Deep Neural Network models: \seqseq{} and \transformer{}. Both architectures take as input the set of normalised instructions that constitute a functions and output a prediction of its name token by token.

\vspace{-0.3cm}
\subsection{Instruction Normalisation}
All our architectures take as input the sequence of normalised assembly instructions. Instructions are normalised with the purpose of reducing their total number  by removing mostly unnecessary information. We follow a normalisation process similar to the one proposed in \cite{SAFE}: 
we replace all base memory addresses with the special symbol \texttt{ MEM} and all immediates whose absolute value is above some threshold (we use $5000$ in our experiments) with the special symbol \texttt{IMM}. 
We do so because raw operands are of small benefit; for instance, the displacement given by a jump is useless  (e.g., instructions do not carry with them their memory address), and, it may worsen performances by artificially inflating the number of different instructions.
In our normalisation the instruction \texttt{mov EBX,}$6000$ becomes \texttt{mov EBX,IMM}, \texttt{mov EBX,}$[0$\texttt{x}$3435423]$  becomes \texttt{mov EBX,MEM},
while the instruction \texttt{mov EAX,}$[$\texttt{EBP}$-8]$ is not modified. 
\vspace{-0.3cm}
\subsection{Sequence Transduction Models} \label{sec:seq}

Sequence Transduction Models are usually used to solve NLP problems such as Neural Machine Translation (NMT). These models take as input a sequence terms and output a transducted sequence. These architectures are natural candidates for our problem due to its similarity with the translation task; in our task we are essentially translating from assembly code to small sets of tokens in human language.
Generally, NMT models are composed by an \emph{encoder} that takes the input sequences and return a set of statuses $c$ and a \emph{decoder} that takes $c$ and output the probability of an output sequence $Y$: 
\begin{equation}
	p(Y) = \prod_{t=1}^{T} p(y(t) | {y_i, ..., y_{t-1}}, c))
\end{equation}

In NMT-like tasks, to generate the output sequence, at each time step $t$ the model outputs a probability distribution over the output vocabulary.

\subsubsection{\seqseq{} Model.}
We use the \seqseq{} architecture proposed in \cite{bahdanau2014neural}: the encoder consists of a bidirectional RNN with Long-Short Term Memory (LSTM) cells. The use of a bidirectional encoder is important since it allows to compute for each instruction an hidden state vector that takes into account the instruction itself and its previous and following context. The decoder is a forward RNN connected to the encoder by an attention mechanism that allows to better model long distance dependencies which represent a critical aspect in assembly code. 

\paragraph{Parameters for \seqseq{} model.} We used an embedding size of 256 for the input and output tokens. For the encoder we used a two layers bidirectional RNN with hidden state size of 256. Equally for the decoder we used a unidirectional RNN with hidden state size of 256. Encoder and Decoder were connected using Bahdanau attention \cite{bahdanau2014neural}. The total number of parameters for the \seqseq{} model is 52618793. The inference step uses a beam-search strategy with beam size 1. We used the implementation provided by OpenNMT-py \footnote{https://github.com/OpenNMT/OpenNMT-py}.

\subsubsection{Transformer.} The \transformer \cite{transformer} is an encoder-decoder architecture entirely based on attention mechanism. The network consists of a set of $N$ stacked encoders and a set of $N$ stacked decoders. The encoder is composed of a stack of $N$ identical layers. The bottom-most layer is fed with the embedding vectors of the input sequence, whereas all the other layers are fed with the output of the previous encoder. Each layer consists of two sub-layers: a multi-head self-attention mechanism used to understand which are the relevant tokens in the input sequence and a fully connected feed-forward network independently applied to each position. In the same way, the decoder is composed of a stack of $N$ identical layers. Each layer consists of three sub-layers: a masked multi-headed mechanism over the decoder input, another multi head attention over the encoder stack output and a final feed forward layer.

\paragraph{Parameters for \transformer{} model.} We used an embedding size of 256 for the input and output tokens. For the encoder we used 6 encoding layers with 8 heads of attention, hidden state size of 256 and hidden feed forward size of 2048. We used the same values for the decoder. The total number of parameters for this model is 67482921. The inference step uses a beam-search strategy with beam size 1. We used the implementation provided by OpenNMT-py.

%% file: sec_6_exp_result.tex

\section{Evaluation}
\label{sec:experiments}

In this section we describe our tests on non-malware binaries. Our evaluation is divided in three main tests. The first experiment is used to train our architectures and select the best performing one. 
In the other two tests we investigate the performance of our best performing architecture on specific applications, and we test the benefit of using fine-tuning to increase its performance. 
\vspace{-0.4cm}
\subsubsection{Fine-tuning.} This is a really popular technique in NLP \cite{fine-tuning}, that proved extremely effective when applied on models that are derived from the Transformer and trained on extremely large general purpose datasets \cite{BERT}.
The main idea is that a model trained on a large general purpose dataset will learn general related patterns with a given distribution, in our case the ones between assembly code and natural language words. However, specific tasks may be characterized by small datasets with differet distributions. Fine-tuning the pre-trained model on such smaller dataset often leads to high performance tailored models for the specific task at hand. 
\vspace{-0.4cm}
 \subsubsection{Tests roadmap.} In details the test are: 
\begin{itemize}
\item {\bf \ubuntudataset{} Training and Test:} we train and test the \seqseq{} and \transformer{} architectures using the \ubuntudataset{}. Moreover, on a subset of \ubuntudataset{} we compare the Transformer with \debin{}.   
The best performing model of this test is the \transformer{}. In the other tests we will investigate how the \transformer{} trained on \ubuntudataset{}, namely \transformerPT{}, performs on specific tasks, with and without fine-tuning.
 
\item  {\bf MultiOpt Test:} we evaluate the performance of \transformerPT{}  on a dataset composed by several applications compiled with 4 different optimisation levels -O($0-4$) and several compilers. Specifically, we test if the fine-tuning on a certain optimisation level increases the performance on the other levels. 

\item  {\bf SameDomain Test:}  we test if the fine-tuning performed on a specific application, namely \textit{busybox}, increases the performance on a different but similar application, namely the \textit{coreutils} binaries. The idea is to see if re-training the \transformerPT{} on an application for a certain subdomain (in this case system utilities) increases its performance on an application of the same domain.  

\end{itemize}

\subsubsection{Evaluation Metrics.}\label{sec:mecmetrics}
Following \cite{NERO}, we use as evaluation metrics the classical precision ($P$), recall ($R$) and f1-score ($F1$). More precisely, given the set of vocabulary words in the actual name ${x}: \{{x_1},{x_2},...,{x_n}\}$,  and the set of tokens in the prediction $\hat{x}:\{\hat{x}_1,\hat{x}_2,..., \hat{x}_n\}$, we define a membership function as:
\begin{equation}
\alpha(\hat{x}_i, x) = 
\left\{ 
\begin{array}{ll} 1 \; \text{if} \; \hat{x}_i \in {x}\\
0 \; \text{else} \\
\end{array}
\right.
\label{eq:alpha}
\end{equation}
and we compute precision, recall and f1-score as:
\begin{multicols}{3}
	\noindent
	\begin{equation}
	P = \frac{\sum_i{\alpha(\hat{x}_i, {x})}}{|\hat{x}|}
	\end{equation}
	\begin{equation}
	R = \frac{\sum_i{\alpha(\hat{x}_i, {x})}}{|{x}|}
	\end{equation}
	\begin{equation}
	F1 = 2 \frac{P+R}{P*R}
	\end{equation}
\end{multicols}

\subsection{\ubuntudataset{}: Training and Test}\label{sec:results}

\subsubsection{Training.} We trained each model on the Train set from  \ubuntudataset{} for a maximum of 30 epochs. We used a batch size of 512, and \emph{Adam} optimizer with decaying learning rate.
After each training epoch we evaluate the performance of the model on the Validation set. We used an early stopping mechanism, stopping the training when the f1-score on the Validation set does not decrease for more than 2 epochs. We took the model with the highest f1-score on the Validation set and tested it on the Test set. 

\vspace{-0.4cm}
\subsubsection{Results.}\label{lb:resultts}
The results are in Table \ref{tb:maintest}; the \transformer{} outperforms \seqseq{} reaching an f1-score of $0.230$ against $0.122$ from \seqseq{}. 
As observed also in \cite{DEBIN}, it is likely that such measures are underestimating the real performance of our system, the reasons have been described in Section \ref{challange:testing}. In our qualitative evaluation on malware (Section \ref{sec:qermw}) we have findings that confirm this hypothesis.  
\vspace{-0.4cm}
\begin{table}[H]
	\centering
	\caption{Results on \ubuntudataset{} Test set} \label{tb:maintest}
		\begin{tabular}{ l | c c c c c c }
			  &  \phantom{x} &  prec. & \phantom{x} &  rec. & \phantom{x} &  f1 \\
			\hline
			\multirow{1}{*}{\seqseq{}} &&  0.174 && 0.095 && 0.122  \\
			\multirow{1}{*}{\transformer{}} &&  \textbf{0.269} && \textbf{0.200} && \textbf{0.230} \\
			\bottomrule
		\end{tabular}

\vspace{-0.4cm}
\end{table}
\vspace{-0.4cm}
\subsubsection{Comparison with \debin{}.}

We compared our solution with \debin{} and RANDOM. We did not compare our solution with NERO since their solution has not been released.  
RANDOM is a basic prediction strategy in which tokens are randomly associated to functions with a probability that respects their frequency in the training set (as an example, ``get" will be sampled more frequently than ``socket").
This kind of strategy has a slight edge on the pure random one. Note that we used \debin{} as released, that is we do not trained their model from scratch on the \ubuntudataset{}. 
 
The comparison has been performed on a subset of our Test set. We took all the binaries contained in our Test set, for each binary we used \debin{} to get a table of the missing debug symbols. We discarded all binaries for which \debin{} was crashing or taking more than 30 minutes to analyse. We then took all predictions for functions that were not in the symbol table of the original files, this is because \debin{} does not predict symbols that are already present in the symbol table of the binary (e.g., exported functions). 
At the end of this process we obtained $49275$ functions. For each of these functions we took the name predicted by \debin{} and we normalised it using the vocabulary described in Section \ref{sec:normname}, obtaining for each name a set of representative tokens. 

\paragraph{Results.}
Results from this test are in Table \ref{comparison:debing}. 
We can see that all solutions are markedly better than the RANDOM strategy. On this dataset the  \transformer{} outperforms the other solutions reaching an f1-score of $0.194$, beating \debin{} $0.046$ and \seqseq{} $0.108$. 
We note that the relatively low value of \debin{} is not necessarily in contrast with what reported in \cite{DEBIN}; their tests are done for the entire naming category, that includes not only the task of naming functions but also the one of naming variables.  
Variable names reuse common patterns (as an example $fd, file, f$ are standard variable names for files), and this could justify an higher rating than the one obtained by looking only at their performances on function names.

\begin{table}[H]
	\centering
	
\caption{Results and comparison with \debin{} on \debin{}-Test subset containing $48k$ functions.}\label{comparison:debing}
	\begin{tabular}{ l | c c c c c c }
		 &  \phantom{x} &  Prec. & \phantom{x} &  Rec. & \phantom{x} &  f1 \\
		\hline
		RANDOM                      && 0.008 && 0.008 && 0.008  \\

		\debin{}       		         && 0.047 && 0.046 && 0.046  \\
		\multirow{1}{*}{\seqseq{}}    && 0.144 && 0.086 && 0.108  \\
		\multirow{1}{*}{\transformer{}} && \textbf{0.229} && \textbf{0.176} && \textbf{0.199}  \\
		\bottomrule
	\end{tabular}
\end{table}

\vspace{-1cm}
\subsection{MultiOpt Test}\label{sec:mcompilers}
\vspace{-0.5cm}
\begin{table}[H]
	\centering
		\caption{Results on MultipleCompilers Dataset for MultiOpt test.} \label{tb:multcompdataset}
	\begin{tabular}{ l | c c c c c c }
		& \phantom{x} &  O$1$ - f1 & \phantom{x} &  O$2$ - f1 & \phantom{x} &  O$3$ - f1 \\
		\hline
		\transformerPT{} (w/o Fine-tuning) &&  0.134 && 0.145 && 0.142 \\
		\transformer{} Fine-tuned && \textbf{0.261} && \textbf{0.275} && \textbf{0.276} \\
		\transformer{} Scratch    &&  0.104 && 0.094 && 0.093 \\
		\bottomrule
	\end{tabular}

\end{table}
\vspace{-0.3cm}
\subsubsection{MultipleCompilers Dataset.} We used the dataset of \cite{BAR19} that contains different packages compiled with 9 different compilers\footnote{clang-3.8, clang-3.9, clang-4.0, clang-5.0, gcc-3.4, gcc-4.7, gcc-4.8, gcc-4.9, gcc-5} and optimization levels from O$0$ to O$3$. On this dataset we selected three packages: \textit{binutils-2.30}, \textit{coreutils-2.29}, \textit{curl-7.61.0}. We split functions in four folds, one for each optimization level. We used the fold with O$0$ for fine-tuning the pre-trained model. Note that we filtered out duplicated functions: fold O$0$ does not include the functions contained in the other three folds. After the filtering, fold O$0$ contains 76727 functions, fold O$1$ contains 61651 functions, fold O$2$ contains 46842 functions, and fold O$3$ contains 43930 functions. The results for this dataset are reported in Table \ref{tb:multcompdataset}.
\vspace{-0.3cm}
\subsubsection{Results without fine-tuning.} We took the \transformerPT{} pre-trained on \ubuntudataset{} and we used it to predict the names of functions in folds O$1$,O$2$,O$3$. Interestingly, the model shows the best performances on higher optimizations. We argue that this is probably due to the fact that the majority of packages in ubuntu repositories are compiled with higher optimization. Therefore, the pre-trained model performs better on these kind of functions since they contain patterns observed in the training phase. 
\vspace{-0.3cm}
\subsubsection{Results with fine-tuning.} We fine-tune the pre-trained \transformerPT{} for 5 epochs using the functions compiled with optimization O$0$. After the fine-tuning we predict the name for the functions with other optimization levels. The results clearly show the benefits of fine-tuning: the f1-score of the fine-tuned \transformer{} on all the optimization levels is close to 2x the one from the non-fine-tuned model.
\vspace{-0.3cm}
\subsubsection{Results when training from scratch.} Using the O$0$ fold we trained a transformer model from scratch. We use functions with optimization O$1$ as Validation set. As for our pre-trained model we stopped the training after 24 epochs when the f1-score on the Validation set did not grow for more than 2 epochs. The maximum f1-score is $0.104$, that is worse than the minimum achieved by the pre-trained and the fine-tuned model.  This was largely expected, as a model trained from scratch on a smaller dataset only learns a limited number of patterns.

\subsection{SameDomain Test}\label{sec:samedomain}
\vspace{-0.4cm}
\begin{table}[H]
	\centering
	\caption{Results on SameDomain Datset for SameDomain Test. }
	\begin{tabular}{ l | c c c c c c }
		& \phantom{x} &  Prec. & \phantom{x} &  Rec. & \phantom{x} &  f1 \\
		\hline
		\transformerPT{} (w/o Fine-tuning)  &&  0.200 && 0.166 && 0.181 \\
		\transformer{} Fine-tuned && \textbf{0.240} && \textbf{0.202} && \textbf{0.220} \\
		\transformer{} Scratch    && 0  && 0 && 0 \\
		\bottomrule
	\end{tabular}
\label{tb:samedomaindataset}
\end{table}

\vspace{-0.35cm}
\subsubsection{SameDomain Dataset.}
We compiled \textit{busybox-1.31.1}  with gcc 7.4.0 and all four optimisation levels. From these binaries we obtained 11897 functions that will constitute our fine-tuning Train set.
The Test set is composed by \textit{coreutils-2.29} compiled with 9 different compilers and optimization levels from O$0$ to O$3$, after processing the functions we obtained 60770 samples. 
\vspace{-0.35cm}
\subsubsection{Results without fine-tuning.}  Without fine-tuning \transformerPT{} has an f1-score of $0.18$. Such result is lower than the one obtained on the Test set of \ubuntudataset{}. We believe that is due to the fact that this dataset contains multiple different compiler optimizations, including the O$0$ that is not frequently used for pre-compiled ubuntu packages (see also similar explanation for tests in previous section).
\vspace{-0.35cm}
\subsubsection{Results with fine-tuning.} We fine-tuned the model re-training it for $5$ epochs. 
With fine-tuning we got an overall increment of $4$ points, the f1-score reaches the value of $0.22$. This corresponds to a relative improvement of performances of around $22\%$ with respect to the pre-trained model. 
 This confirms that fine-tuning is a good strategy when the naming has to be performed on applications of a certain specific domain. 
\vspace{-0.35cm}
\subsubsection{Results when training from scratch.} 
When the model is trained from scratch we reach a f1-score of $0$. This is because the model incorrectly learns to associate any function to an empty name. This is probably due to the fact that the training set is too small to learn any meaningful  relationships.

%% file: sec_7_human.tex
\section{Test on Malware}\label{sec:malware}
We tested the  Transformer architecture on real world malware. We performed a quantitative evaluation on linux malware obtained from Virus-share (Section \ref{sec:malquant}), and a qualitative analysis on two malware for which the source code is available (Section \ref{sec:qermw}). 

\vspace{-0.35cm}
\subsection{Quantitative Evaluation}\label{sec:malquant}
In this section we describe our quantitative evaluation on malware.

\begin{table}[H]
	\centering
	\caption{Results on \malwaredataset. \textit{singleton} represents all samples that where not labelled by\textit{AVCLASS}. \textit{others} groups the families containing only one sample.}
		\resizebox{\textwidth}{!}{\begin{tabular}{ l | c c c c c c c c c c c c c c c c c c c c c c c c c c }
			& \phantom{x} & chinaz  & \phantom{x} & dnsamp & \phantom{x} & drtycow  & \phantom{x} &  gafgyt & \phantom{x} & intfour & \phantom{x} &  ladvix  & \phantom{x} &  mirai & \phantom{x} & snessik & \phantom{x} & sotdas & \phantom{x} & yangji  & \phantom{x} & znaich & \phantom{x} & singleton & \phantom{x} & others\\
			\hline
			Number of samples && 4 && 12 && 5 && 217 && 3 && 2 && 17 && 3 && 4 && 3 && 7 && 57 && 11 \\
			\hline
			\transformerPT{} (w/o Fine Tuning) && 0.119 && 0.122  && 0.282 && 0.156 && 0.404 && 0.356 && 0.124 && 0.302 && 0.105 && 0.141 && 0.119 && 0.221 && 0.103 \\
			\transformer{} Fine Tuned  && \textbf{0.551} &&  \textbf{0.463} && \textbf{0.582} && \textbf{0.914} && \textbf{0.864} && \textbf{0.667} && \textbf{0.577} && 	\textbf{0.857} && \textbf{0.492} && \textbf{0.933} && \textbf{0.581}  && \textbf{0.435} && \textbf{0.401} \\
			\transformer{} Scratch   && 0.401  && 0.331  && 0.560 && 0.870 && 0.772 && 0.571 && 0.455 && 0.857 && 0.346 && 0.910 && 0.440 && 0.194 && 0.280 \\
			\bottomrule
	\end{tabular}}
 \label{tb:maldataset}
\end{table}

\vspace{-0.7cm}
\subsubsection{\malwaredataset.} We built a dataset of malware downloading an ELF collection from Virus-share \footnote{\url{https://tracker.virusshare.com:7000/torrents/VirusShare_ELF_20190212.zip.REDACTED}}. To have a reliable ground-truth  we only considered malware that are not stripped. Moreover, we selected only the malware for AMD64. We used VirusTotal\footnote{www.virustotal.com} to obtain the antivirus labels for each selected sample. We used \textit{AVCLASS} \cite{avclass} to assign a family to each sample. At the end of this process we obtained 406 malware belonging to 23 families. The families are very unbalanced, the majorities of samples ($217$) belongs to the class \textit{gafgyt}, 11 families contains only one sample, and 57 samples were classified from \textit{AVCLASS} as SINGLETON. We disassembled all samples with IDA Pro and we filtered function names using the procedure and the vocabulary described in Section \ref{sec:dataset}. In total, we gathered 156316 functions.
\vspace{-0.3cm}
\subsubsection{Test Description and Dataset Split.} We will perform three kind of tests: one taking the Transformer architecture trained on the \ubuntudataset{} (namely \transformerPT{}) as it is; another in which we {\em fine-tune} the trained architecture by performing a small re-train on a Training set that we will describe later; and, a final one where we used the same aforementioned Training set to train a Transformer model from scratch. The Training set contains the 61 samples of \textit{tsunami} family, the remaining samples are used for Test set. We decided to use a rather limited Training set to model a worst case scenario where there are only few malware that can used to build a labeled dataset.
\vspace{-0.3cm}
\subsubsection{Results.}
The results of our tests are in Table \ref{tb:maldataset}. 

\paragraph{Tests without fine-tuning.} We tested \transformerPT{} on the \malwaredataset{}. The average f1-score for all the classes is 0.196 and it is lower than the one on the \ubuntudataset{}  (f1-score of 0.230). However, as we already mentioned this f1-score is possibly underestimating the real performance by predicting names that are meaningful but not exactly equal to the ones in the dataset. This hypothesis is confirmed by our qualitative analysis in  Section \ref{sec:qermw}.

\paragraph{Tests with fine-tuning.} We fine-tuned the \transformerPT{} model already trained on \ubuntudataset{} using \malwaredataset{} Training set. We stopped the retrain after 43 epochs when the performance on the Training set where not improving any more. The fine tuned model clearly shows the benefits of a domain specific fine-tune. We reached an f1-score of 0.640. We think that during the fine-tuning the network learns the specific domain, that is constituted mainly by encryption functions, network functions,  IO operations, functions that gather informations from the OS and statically linked libraries.

\paragraph{Training the model from scratch.} We also used the same Training set to train a Transformer from scratch. Interestingly, this model reaches an high f1-score. We argue that this is due to the code reuse between different families and static linking.
As a matter of fact while the Transfromer trained from scratch achieve good performance on certain families (\textit{gafgyt} \textit{intfour},\textit{snessik}), it performs markedly worse, with respect to the fine-tuned model, on others.
As an example on the singleton group the from scratch model reaches an f1-score of 0.194 (vs. pre-trained transformer 0.221), while the fine tuned model reaches an f1-score of 0.401.

\subsection{Qualitative Evaluation on Malware}
\label{sec:qermw} 

We performed a qualitative evaluation on the infamous botnet MIRAI, for which source code has been leaked \footnote{\url{https://github.com/jgamblin/Mirai-Source-Code}}, and on the educational ransomware gonnacry \footnote{\url{https://github.com/tarcisio-marinho/GonnaCry}}. 
We decided to use malware for which the source code was available in order to manually asses each prediction by looking also at the original source code.  Our hypothesis is that the f1-score is underestimating the performance of the pre-trained \transformerPT{}.
We will show that the prediction of our model can provide to a reverse engineer some useful insights on a stripped binary.

\input{mirai_prediction.tex}

\subsubsection{Analysis of Predictions on the Mirai Botnet.}

We compiled the botnet from the source code with optmization O$0$ and gcc-7.4. We disassembled it using IDA Pro obtaining 74 functions. 
We report in Table \ref{comparisonmirai} the comparison of reference, prediction of \transformerPT{} and DEBIN. 
The value of the f1-score computed on such predictions is 0.157 for\transformerPT{} and 0.013 for Debin. We identified the following interesting aspects:

\begin{itemize}
\item {\bf Networking and Checksum functions}: all networking functions are associated with a set of tokens related to network functionalities, that is usually composed by {\sf send}, {\sf send to} or {\sf send request}, we highlight {\sf teardown\_connection} (row 35) that is correctly named as {\sf closed}. The function {\sf checksum\_generic} (row 23) is correctly named as {\sf crc}. 

\item {\bf String and memory related}: The function {\sf mem\_exists} (row 8) looks for a string in a buffer, and it is named {\sf str cmp}, another example is {\sf util\_memseach} (row 60) that is named as {\sf find} and it also finds a string in memory.  Functions that check if a string is in a particular format or if a character is in a particular format are almost correctly predicted (e.g. {\sf util\_isupper} (row 64) is named in {\sf is low}).

\item {\bf File Operations}: The function {\sf killer\_kill\_by\_port} (row 27) finds the process listening on a certain port by reading from the proc filesystem. In this case, our model predicts {\sf process read dir} that correctly represent part its behaviour. The function {\sf has\_exe\_access} (row 28) opens the file {\em /proc/[pid]/exe}, in our model the functions is named as {\sf read pid file}. The function {\em memory\_scan\_match} (row 29) search for a certain string in a file, our model predicts {\sf get from file}.

\item {\bf Other functions}:  The function {\sf attack\_kill\_all} (row 2) is translated in {\sf kill all}, by looking at the code the function is iterating on an array of PIDs and killing each process in the array. The function {\sf attack\_start} (row 4) is starting a set of child processes, and it is correctly identified as {\sf run child}.  In {\sf attack\_get\_opt\_str} (row 5) a value is searched in an array, and this is correctly named as {\sf find}. {\sf attack\_get\_opt\_ip} (row 7) returns an ip address and this is named as {\sf addr} by our network. The function {\sf add\_attack} (row 8) performs a reallocation of an array incrementing its size and it adds an element, the network predicted name {\sf add} fits this behaviour. 
\end{itemize}

\subsubsection{Analysis of Predictions of Gonnacry.}

\begin{table}[h]
\scriptsize
\caption{Prediction of the Transformer and DEBIN on gonnacry}

\begin{tabular}{lllll} 
Row \# & Reference &  Transformer Prediction & Debin & Souce code analysis \\ \hline
1 & main & write & load\_init\_file &  \\
2 & find\_files & scan dir & listdir & find all files in a given directory \\
3 & create\_files\_desktop & operator  & blurb & wrapper for other functions \\
4 & save\_into\_file\_files\_list & write file & save\_session & write files \\
5 & save\_into\_file\_encrypted\_list & write file  & make\_openvpn\_gui\_conf & write file \\
6 & read\_from\_file\_encrypted\_files & load & readRuleFile & read from a file and parse its content  \\
7 & get\_filename\_ext & get file extension  & \_\_stack\_chk\_fail\_local & get extension of a file \\
8 & get\_home\_enviroment & get home dir  & init\_file\_name & get home path by manipulating string \\
9 & get\_username & get user name  & archdep\_default\_rtc\_file\_name & return the logged username \\
10 & get\_trash\_path & new  & set\_player\_name &  allocate the memory location for a string and manipulate it\\
11 & get\_media\_path & make path  & get\_and\_append\_filenames & get the path to media by manipulating string \\
12 & get\_desktop\_enviroment & get path  & ccParseRule & return the Desktop path by manipulating string \\
13 & get\_test\_path & strdup  & oldgaa\_strcopy & manipulate a string \\
14 & is\_path & is dir  & ioutil\_opendir & check if a directory exist \\
15 & generate\_key & random  & encrypt\_init & generate a random string \\
16 & append & add  & openfile & append element to a linked list \\
17 & destro & free & free\_screenhack\_list & free the memory of a linked list \\
18 & print & print  & do\_baro & print values in a list \\
19 & length & get  & globus\_i\_ftp\_client\_plugin\_notify\_utime & get a value representing list len \\
20 & encrypt\_files & open  & main & open some files \\
21 & decrypt\_files & open file & CDE\_begin\_execve & open some files \\
22 & shred & write file & connect\_options & overwrite a file with zeros \\
23 & encrypt & copy file  & build & encrypt the content from file and save on another \\
24 & decrypt & copy file  & wav\_merge\_files & encrypt the content from file and save on another \\
\end{tabular}
\label{comparisoncry}
\end{table}

We report in Table \ref{comparisoncry} the comparison of reference, prediction of \transformerPT{} and DEBIN.
The value of f1-score for DEBIN predictions is $0.0825$ while the transformer reaches an f1-score of $0.260$.
On this malware the predictions are rather good and even when they differ from the ground truth, they still express sub-behaviours implemented in the function. 
{\sf get\_desktop\_environment} (row 12) is predicted as {\sf get path}, interestingly, such function returns the desktop path. 
Other interesting examples are the functions {\sf decrypt}/{\sf encrypt} (rows 23, 24),  what they do is to open a file copying its encrypted/decrypted contend in another file,
the Transformer names such functions as {\sf copy file}. The function {\sf generate\_key} (row 15) creates a random string and it is named as {\sf random}.

%% file: mirai_prediction.tex
\begin{table}[]
\scriptsize
\caption{Prediction of the Transformer and DEBIN on mirai}
\begin{tabular}{ l l l l l l l l l }
			 Row \# 	&&   Reference & \phantom{x}  & Transformer Prediction & \phantom{x} & Debin Prediction & \phantom{x} & Description \\  
			\hline
			1	&&	attack\_init	&&	emit load	&&	tnt\_list\_at	&&	call another functions multiple times \\
			2	&&	attack\_kill\_all	&&	kill all	&&	kill\_faio	&&	read process id from array and kill process\\
			3	&&	attack\_parse	&&	set	&&	globus\_gsi\_cert\_utils\_get\_x509\_name	&&	parse a command packet\\
			4	&&	attack\_start	&&	run child	&&	jed\_fork\_monitor	&&	start processes using fork\\
			5	&&	attack\_get\_opt\_str	&&	find	&&	err\_ssl	&&	find a string in an hashmap\\
			6	&&	attack\_get\_opt\_int	&&	get	&&	\_\_fprintf\_chk	&&	get an integer in a dictionary \\
			7	&&	attack\_get\_opt\_ip	&&	addr	&&	shutdown	&&	get an ip address from a dictionary \\
			8	&&	add\_attack	&&	add	&&	tnt\_tuple\_add	&&	reallocate more space for an array and adds an element\\
			9	&&	free\_opts	&&	free	&&	get\_mod\_mask\_for	&&	executes several frees \\
			10	&&	attack\_app\_proxy	&&	operator	&&	globus\_i\_ftp\_client\_plugin\_notify\_symlink	&&	flooding packets\\
			11	&&	attack\_app\_http	&&	sec key	&&	main	&&	flooding packets\\
			12	&&	attack\_app\_cfnull	&&	send request	&&	mitm\_child	&&	flooding packets\\
			13	&&	attack\_gre\_ip	&&	send	&&	Multicast\_receive\_socket	&&	flooding packets\\
			14	&&	attack\_gre\_eth	&&	send request	&&	Multicast\_send\_socket	&&	flooding packets\\
			15	&&	attack\_tcp\_syn	&&	send request	&&	Unicast\_send\_socket	&&	flooding packets\\
			16	&&	attack\_tcp\_ack	&&	send	&&	sendfile\_tcp\_stream	&&	flooding packets\\
			17	&&	attack\_tcp\_stomp	&&	send	&&	query\_mndp	&&	flooding packets\\
			18	&&	attack\_udp\_generic	&&	send request	&&	Unicast\_receive\_socket	&&	flooding packets\\
			19	&&	attack\_udp\_vse	&&	send	&&	InitNetwork	&&	flooding packets\\
			20	&&	attack\_udp\_dns	&&	send request	&&	openSocket	&&	flooding packets \\ 
			21	&&	attack\_udp\_plain	&&	send	&&	establish\_control\_internal	&&	flooding packets \\
			22	&&	get\_dns\_resolver	&&	read	&&	do\_verify\_password	&&	read from the table the dns address or use a random dns \\
			23	&&	checksum\_generic	&&	crc	&&	fwrite	&&	compute a generic checksum function \\ 
			24	&&	checksum\_tcpudp	&&	cmp	&&	LogError	&&	compute checksum for tcp upd packets \\
			25	&&	killer\_init	&&	send to	&&	OpenSocket	&&	open a socket \\
			26	&&	killer\_kill	&&	kill process	&&	jed\_lock\_file	&&	kill a process using pid \\
			27	&&	killer\_kill\_by\_port	&&	process read dir	&&	create\_lockspace	&& finds a process listening on a port by reading on the proc fs \\
			28	&&	has\_exe\_access	&&	read pid file	&&	tty\_create\_lock	&&	try to open /proc/[pid]/exe \\
			29	&&	memory\_scan\_match	&&	get from file	&&	blacklisted\_key\_in\_file	&&	read data from file and check for matches \\
			30	&&	mem\_exists	&&	str cmp	&&	debug	&&	check if a string is contained in a buffer \\
			31	&&	main	&&	handle connection	&&		&&	handle a connection with sockets \\
			32	&&	anti\_gdb\_entry	&&	code expr new	&&	display\_option\_dialog\_popup	&&	assign to a pointer the location of a function \\
			33	&&	resolve\_cnc\_addr	&&	get hash	&&	do\_read\_file	&&	dns query to resolve c\&c server \\
			34	&&	establish\_connection	&&	socket connect	&&	connect\_proxy	&&	create a connection with sockets \\
			35	&&	teardown\_connection	&&	close	&&	pw\_extauth\_check	&&	close a connection \\
			36	&&	ensure\_single\_instance	&&	socket	&&	ntpdatemain	&&	check for a single connection \\
			37	&&	unlock\_tbl\_if\_nodebug	&&	format	&&	dStrHexStr	&&	dynamic call to init table \\
			38	&&	rand\_init	&&	hash	&&	image\_loaded\_cb	&&	initialize random generator \\
			39	&&	rand\_next	&&	key hash	&&	perror	&&	generate a random number \\
			40	&&	rand\_str	&&	str	&&	freeaddrinfo	&&	generate a random string \\ 
			41	&&	rand\_alphastr	&&	encode	&&	SocketClient\_create	&&	generate a random string  \\
			42	&&	resolv\_domain\_to\_hostname	&&	get	&&	mnt\_closeclnt	&&	parse a domain to find subdomain \\
			43	&&	resolv\_skip\_name	&&	utf decode	&&	nfs\_call\_umount	&&	array search \\
			44	&&	resolv\_lookup	&&	send	&&	nfsmount	&&	make a dns query, send and receive data with socket \\
			45	&&	resolv\_entries\_free	&&	free	&&	lsh\_copy\_file	&&	free memory \\
			46	&&	table\_init	&&	gen cmd isra	&&	display\_option\_dialog\_apply	&&	initialize table \\
			47	&&	table\_unlock\_val	&&	code expr new	&&	lsh\_get\_cstring	&&	unlock a location in a table \\
			48	&&	table\_lock\_val	&&	code expr new	&&	lsh\_object\_alloc	&&	lock a location in a table \\
			49	&&	table\_retrieve\_val	&&	get	&&	werror	&&	get a value from a table \\
			50	&&	add\_entry	&&	make	&&	werror\_write\_raw	&&	add a new entry in a table \\
			51	&&	toggle\_obf	&&	lib image rgb to rgb	&&	verbose	&&	obfuscate with xor a location in the table \\
			52	&&	util\_strlen	&&	str len	&&	strlen	&&	compute the len of a string \\
			53	&&	util\_strncmp	&&	compare	&&	parse\_dir	&&	check if 2 strings are equals \\
			54	&&	util\_strcmp	&&	operator	&&	smatch	&&	check if 2 strings are equals \\
			55	&&	util\_strcpy	&&	add	&&	buf\_strdup	&&	copy strings \\
			56	&&	util\_memcpy	&&	put	&&	memcpy	&&	copy a portion of the memory \\
			57	&&	util\_zero	&&	str trim	&&	\_\_memcpy\_chk	&&	set a memory portion to zero \\
			58	&&	util\_atoi	&&	py byte	&&	\_\_stack\_chk\_fail\_local	&&	convert a numerical string to an integer \\
			59	&&	util\_itoa	&&	date time to time	&&	write	&&	convert an integer into a string \\
			60	&&	util\_memsearch	&&	find	&&	game\_set\_options\_from\_defaults	&&	find a string in a buffer \\
			61	&&	util\_stristr	&&	find char	&&	lsh\_string\_data	&&	find for a substring \\
			62	&&	util\_local\_addr	&&	connect	&&	nfs\_callback\_address	&&	open a connection and return an address \\
			63	&&	util\_fdgets	&&	get	&&	open\_control\_device	&&	get data from file \\ 
			64	&&	util\_isupper	&&	is low	&&	globus\_i\_ftp\_client\_plugin\_notify\_utime	&&	check if a character is upper case \\
			65	&&	util\_isalpha	&&	is low case	&&	globus\_ftp\_client\_pl ugin\_set\_connect\_func	&& 	check if a character (low. and up. case) is alphabetic \\
			66	&&	util\_isspace	&&	is space	&&	globus\_ftp\_client\_operationattr\_get\_dcau	&&	check if a character is a  space \\
			67	&&	util\_isdigit	&&	is digit	&&	globus\_i\_ftp\_client\_feature\_set	&&	check if a character is a  digit \\
		\end{tabular}
	\label{comparisonmirai}
\end{table}

%% file: sec_8_conclusion.tex

\section{Conclusion}
\label{sec:conclusion}

This paper proposes a study on the problem of predicting names of functions in stripped binary code. We tested state-of-the-art solutions in machine translation finding a rather good carryover on our problem. L
Moreover, we created a large public dataset of functions that can be used to further the research on the topic. The results that we found are encouraging, and pave the way for further studies. 
We believe that many improvements are possible, and that one of the main challenges is to find faithful metrics that would capture the performances perceived by a human. 
To this end it would be beneficial to investigate the following directions:
\begin{itemize}
\item {\bf Multi-references dataset}:  an idea is to create a dataset where each function is associated with multiple reference names. This would ameliorate the evaluation problem since it would take into account the possibility of deviating from a single canonical name. 
\item  {\bf Extensive human evaluation}: another line is to perform an extensive investigation using humans. It would be beneficial to evaluate this kind of solutions using experienced programmers that look at the predicted name and at the source code of the function.

\item  {\bf Metrics based on NLP}: finally, it would be interesting to use metrics based on NLP. Recently, BERTScore \cite{bert-score} proposes new metrics based on contextual word embedding computed with BERT \cite{BERT}. The intuition is that this process takes into account the semantic of the words in the predicted and reference sentences. 
\end{itemize}

{\bf Acknowledgment:} This work was partially supported by Giuseppe Di Luna's Axa Postdoctoral Fellowship and by Sapienza University of Rome's project RM11916B75A3293D.